%% file: main.tex
\documentclass[sigconf]{acmart}
\AtBeginDocument{%
  }

\usepackage{algorithm}
\usepackage{algpseudocode}
\usepackage{enumitem}
\setlist[itemize]{leftmargin=*, nosep}
\usepackage{multirow} 
\usepackage{array}
\usepackage{float}

\usepackage{dblfloatfix}
\usepackage{placeins}
\usepackage{graphicx}   
\usepackage{booktabs}   
\usepackage{multirow}   
\usepackage{color, xspace}

\newcommand{\name}{FedDis\xspace}

\newcommand{\ie}{\emph{i.e.,}\xspace}

\begin{document}

\copyrightyear{2026}
\acmYear{2026}
\setcopyright{cc}
\setcctype{by}
\acmConference[WWW '26]{Proceedings of the ACM Web Conference 2026}{April 13--17, 2026}{Dubai, United Arab Emirates}
\acmBooktitle{Proceedings of the ACM Web Conference 2026 (WWW '26), April 13--17, 2026, Dubai, United Arab Emirates}
\acmPrice{}
\acmDOI{10.1145/3774904.3792663}
\acmISBN{979-8-4007-2307-0/2026/04}

\title{The Name of the Title Is Hope}

\author{Chengyang Zhou}
\orcid{0009-0004-8572-5429}
\authornote{Also with the Key Laboratory of Symbolic Computation and Knowledge Engineering of Ministry of Education, Jilin University.}
\affiliation{
  \institution{Jilin University}
  \city{Changchun}
  \country{China}
}
\email{chengyang25@mails.jlu.edu.cn	}

\author{Zijian Zhang}
\orcid{0000-0003-1194-8334}
\authornotemark[1]
\authornote{Corresponding authors.}
\affiliation{
  \institution{Jilin University}
  \city{Changchun}
  \country{China}
}
\email{zhangzijian@jlu.edu.cn}

\author{Chunxu Zhang}
\orcid{0000-0003-0825-872X}
\authornotemark[1]
\affiliation{
 \institution{Jilin University}
 \city{Changchun}
 \country{China}
}
\email{zhangchunxu@jlu.edu.cn}

\author{Hao Miao}
\orcid{0000-0001-9346-7133}
\affiliation{
  \institution{The Hong Kong Polytechnic University}
  \city{Hong Kong}
  \country{China}}
\email{hao.miao@polyu.edu.hk}

\author{Yulin Zhang}
\orcid{0009-0003-3848-3175}
\affiliation{
  \institution{Jilin University}
  \city{Changchun}
  \country{China}
}
\email{ylzhang2123@mails.jlu.edu.cn	}

\author{Kedi Lyu}
\orcid{0000-0002-7905-3759}
\affiliation{
  \institution{Jilin University}
  \city{Changchun}
  \country{China}
}
\email{kdlyu@jlu.edu.cn}

\author{Juncheng Hu}
\orcid{0000-0002-6232-9093}
\authornotemark[2]
\affiliation{
  \institution{Jilin University}
  \city{Changchun}
  \country{China}
}
\email{jchu@jlu.edu.cn}

\renewcommand{\shortauthors}{Chengyang Zhou et al.}

\title{FedDis: A Causal Disentanglement Framework for Federated Traffic Prediction}
% \title{FedDT: A Federated Disentanglement Framework for Traffic Prediction}

\begin{abstract}
\input{sections/1_Abstract}

\end{abstract}




\maketitle

\input{sections/2_Introduce}
\input{sections/4_Method}

\input{sections/5_Experiments}

\input{sections/3_RelatedWorks}
\input{sections/6_Conclusion}
\input{sections/7_Acknowledgements}
\clearpage
\bibliographystyle{ACM-Reference-Format}
\balance
\bibliography{sections/8_References}

\appendix
\input{sections/9_Appendix}
\end{document}

%% file: sections/1_Abstract.tex
Federated learning offers a promising paradigm for privacy-preserving traffic prediction, yet its performance is often challenged by the non-identically and independently distributed (non-IID) nature of decentralized traffic data. 
Existing federated methods frequently struggle with this data heterogeneity, typically entangling globally shared patterns with client-specific local dynamics within a single representation.
In this work, we postulate that this heterogeneity stems from the entanglement of two distinct generative sources: client-specific localized dynamics and cross-client global spatial-temporal patterns.
Motivated by this perspective, we introduce \name, a novel framework that, to the best of our knowledge, is the first to leverage causal disentanglement for federated spatial-temporal prediction. 
Architecturally, \name comprises a dual-branch design wherein a Personalized Bank learns to capture client-specific factors, while a Global Pattern Bank distills common knowledge. 
This separation enables robust cross-client knowledge transfer while preserving high adaptability to unique local environments. 
Crucially, a mutual information minimization objective is employed to enforce informational orthogonality between the two branches, thereby ensuring effective disentanglement. 
Comprehensive experiments conducted on four real-world benchmark datasets demonstrate that \name consistently achieves state-of-the-art performance, promising efficiency, and superior expandability. 
%To ease the reproducibility, we release our implementation code online\footnote{\url{https://github.com/Jlu-zcy/www2026_FedDis}}.

%% file: sections/2_Introduce.tex
\section{Introduction}
The proliferation of location-based services has generated massive traffic data, providing new opportunities to support urban planning and intelligent transportation systems \cite{wang2021libcity, lv2014traffic,fang2025unraveling,wang2023pattern,ma2025bist,ma2025less}, centering on the challenge of modeling complex spatial-temporal dependencies across road networks \cite{liu2023robust}.
The development of deep spatial-temporal models has significantly advanced this capability. 
Early approaches such as DCRNN \cite{li2017diffusion} and GWNet \cite{wu2019graph} leverage diffusion graph convolutions and adaptive graph learning, while more recent transformer-based and hybrid architectures, like PDFormer~\cite{jiang2023pdformer} and D2STGNN~\cite{shao2022decoupled}, have achieved further breakthroughs in long-range temporal modeling and dynamic signal disentanglement.

\begin{figure}[!t]
    \centering
    \includegraphics[width=\linewidth]{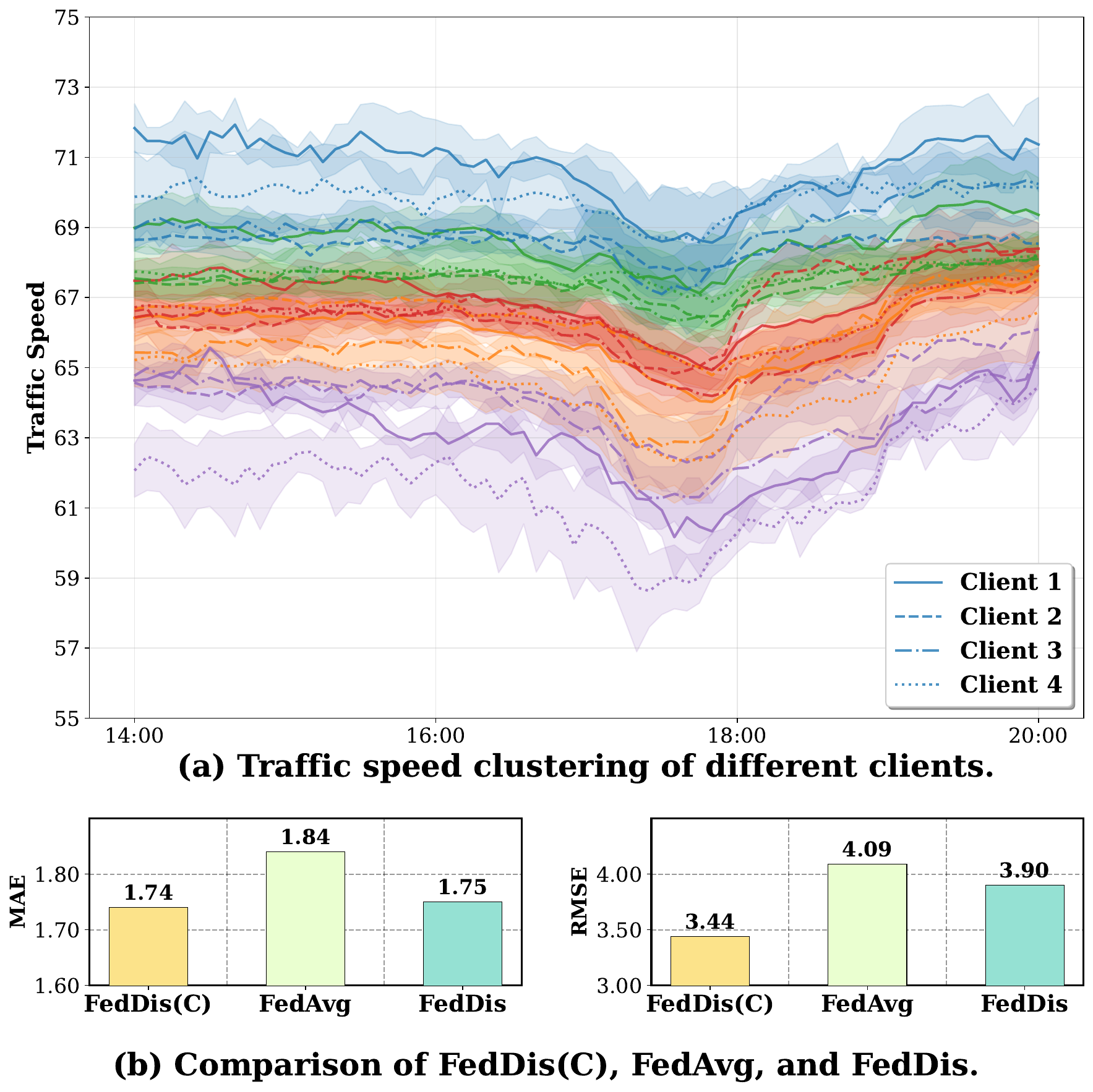} % 自适应当前栏宽
    \caption{Illustration on PEMS-BAY. (a) The illustration of universal spatial-temporal patterns across clients. (b) Performance comparison of centralized, federated, and our \name. }
    \label{fig:flow_cluster}
\end{figure}

However, the effectiveness of these centralized models is predicated on data aggregation, a requirement often unfeasible in practice, where traffic data is siloed across regions due to privacy and ownership concerns. 
Consequently, collaborative prediction without sharing raw data has become a pressing need for transportation authorities. 
Federated Learning (FL)~\cite{mcmahan2017communication, yang2019federated}, a computational paradigm designed to address this challenge, enables cross-client model training while preserving data privacy.

Federated learning methods have emerged for spatial-temporal prediction. FedGRU~\cite{liu2020privacy} models temporal patterns with recurrent networks, CNFGNN~\cite{meng2021cross} captures spatial dependencies via Graph Neural Networks (GNNs), FedGTP~\cite{yang2024fedgtp} exploits inter-client spatial correlations while preserving privacy, SFL~\cite{chen2022personalized} simultaneously learns a global model and client-specific personalized models using relation graphs. Methods such as FedTPS~\cite{zhou2024traffic} and pFedCTP~\cite{zhang2024personalized} address data heterogeneity through pattern sharing or cross-region knowledge transfer.
Despite its potential, applying FL to the spatial-temporal domain reveals a fundamental challenge: learning globally consistent patterns from locally heterogeneous data. 
The non-identically and independently distributed (non-IID) nature of client data not only complicates the extraction of stable global patterns but also makes models highly susceptible to client-specific confounding factors~\cite{liu2023multilevel, liu2025personalized, zhang2025fedstg}. 
Compounding this issue, existing federated frameworks entangle global and local knowledge within a unified representation, failing to explicitly distinguish their roles during optimization~\cite{yang2024fedgtp, meng2021cross, zhang2025fedstg, liu2025personalized, zhang2024personalized}. 
This compromises the model's adaptability to local environments and introduces potential privacy risks.

This challenge is directly reflected in real-world data. As illustrated in Figure~\ref{fig:flow_cluster} (a), each line denotes the mean speed of a traffic pattern cluster, while the corresponding shaded band illustrates its distribution. 
The visualization reveals that different clients exhibit strong commonalities in both their overall speed distributions and their temporal dynamics, highlighting the existence of shared global patterns.
This observation inspires our approach: an effective federated model should first identify common patterns locally, then maintain global insights through cross-client knowledge sharing, all while preserving personalized client dynamics.

To address these challenges, we propose a causal disentanglement framework for federated
spatial-temporal prediction, \name, a dual-branch framework that explicitly disentangles globally stable patterns and personalized local factors. 
The global branch captures stable, cross-client traffic patterns that can be collaboratively aggregated, whereas the personalized branch models client-specific dynamics updated locally. 
This dual-branch causal disentanglement facilitates robust cross-region knowledge transfer, maintains local adaptability, safeguards sensitive client information, and improves interpretability and predictive accuracy.

Empirical results confirm the effectiveness of this design. 
As shown in Figure~\ref{fig:flow_cluster} (b), we compare different training paradigms with our backbone, \ie centralized training (\name (C)), representative federated training (FedAvg), and our disentangled federated training (\name). 
Our \name consistently outperforms {FedAvg}, which jointly optimizes global and local information, achieving lower MAE and RMSE and approaching the performance of its centralized counterpart {\name(C)}. 
These results demonstrate that separating shared and personalized patterns enables more robust, generalizable, and adaptive federated spatial-temporal traffic prediction, effectively addressing both local and global challenges under heterogeneous, decentralized traffic data.

Our major contributions can be summarized as:
\begin{itemize}
    \item We identify and formalize two distinct yet complementary patterns within federated spatial-temporal data: globally stable correlations and personalized local dynamics, and propose a causal disentanglement framework to model them separately, addressing the core challenge of data heterogeneity.
    \item To our knowledge, we propose the first causal disentanglement federated framework for traffic prediction, named \name. It features a novel dual-branch architecture to isolate and learn the two aforementioned components separately, ensuring that client-specific knowledge is not conflated with global patterns.
    \item We conduct extensive experiments on four real-world datasets, demonstrating that \name consistently achieves state-of-the-art performance against competitive baselines. Our in-depth analysis validates that this disentangled approach is robust, scalable, and effective for heterogeneous, decentralized traffic data.
\end{itemize}

%% file: sections/4_Method.tex
\section{Methodology}

\begin{figure*}[!t]
    \centering
    \includegraphics[width=0.8\textwidth]{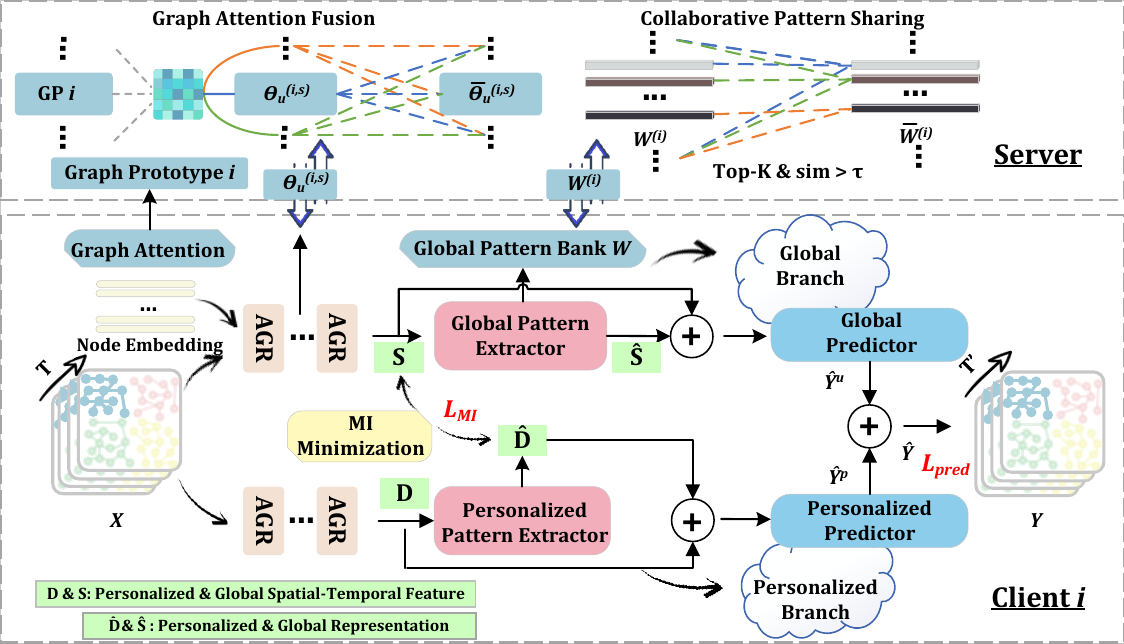}
    \caption{\name framework. Each client adopts a dual-branch architecture with a global branch and a personalized branch, which enables causal decoupling.
    The Global Pattern Extractor maintains a Global Pattern Bank ($\mathbf{W}$), which captures globally shared patterns, while the Personalized Pattern Extractor maintains a Personalized Pattern Bank ($\mathbf{L}$), capturing local dynamic traffic variations.
    Client aggregates node embeddings to yield its Graph Prototypes (GP), which indicate the client's overall structural and properties. On the server side, GP guides the aggregation of sharable parameters $\mathbf{\theta_u^{(i,s)}}$, and the $\mathbf{W}$ facilitates global knowledge sharing across clients.}
    \label{fig:framework}
\end{figure*}

\subsection{Problem Definition}

\noindent\textit{\textbf{Spatial-Temporal Prediction.}}
Let the road network be modeled as a graph $\mathcal{G} = (\mathcal{V}, \mathbf{A})$ with $N=|\mathcal{V}|$ nodes, where each node represents a traffic sensor and $\mathbf{A}\in\mathbb{R}^{N\times N}$ encodes spatial correlations.
Each node $v_i$ is associated with a traffic time series $\mathcal{X}_i = \{x^{t}\}_{t=1}^{T_{\text{total}}}$, where $x^t\in\mathbb{R}^F$ denotes the traffic state at time $t$.
Given historical observations $\mathcal{X}_{t-T+1:t}$, \textbf{spatial-temporal prediction} aims to predict future traffic states over the next $T'$ steps: $\hat{\mathcal{Y}} = \mathcal{F}\big(\mathcal{X}_{t-T+1:t}, \mathbf{A}; \theta \big)$,
where $\mathcal{F}(\cdot)$ denotes a learnable model parameterized with $\theta$, and $ \hat{\mathcal{Y}} \in \mathbb{R}^{N \times T' \times F}$.

\noindent\textit{\textbf{Federated Spatial-Temporal Prediction.}}
Under the privacy constraint that traffic data are partitioned across multiple clients, each client $m$ maintains a local subgraph $\mathcal{G}^{(m)} = (\mathcal{V}^{(m)}, \mathbf{A}^{(m)})$ and its associated traffic observations $\mathcal{X}^{(m)}$. The complete road network $\mathcal{G}$ is formed by the union of all $M$ clients' subgraphs, \ie $\mathcal{G} = \bigcup_{m=1}^{M} \mathcal{G}^{(m)}$. 

In \textbf{federated spatio-temporal prediction}, the goal is to collaboratively learn a global model $\mathcal{F}_{\theta}$ that minimizes the aggregated prediction loss across all clients while preserving data privacy. Formally, the optimization objective is defined as,
\begin{equation}
    \min_{\theta} \sum_{m=1}^{M} w_m \, \mathcal{L}\Big(\mathcal{F}\big(\mathcal{X}^{(m)}_{t-T+1:t}, \mathbf{A}^{(m)}; \theta\big), \mathcal{Y}^{(m)}\Big),
\end{equation}
where $\mathcal{L}(\cdot)$ denotes the local prediction loss for each client, $w_m$ is a weight reflecting the relative importanceof client $m$, $\mathcal{Y}^{(m)}$ represents the future traffic states for client $m$, and $\theta$ are the parameters of the shared global model optimized through federated learning.

\subsection{Framework Overview}
As illustrated in Figure~\ref{fig:framework}, we propose \textbf{\name}, a federated causal disentanglement spatio-temporal prediction framework designed to jointly model personalized and global patterns.

On each client, two stacked Adaptive Graph Convolutional Recurrent (AGR) cells (Section~\ref{subsec:st_feature}) first encode the spatial-temporal data. The feature is then processed by the causal representation decoupling framework (Section~\ref{subsec:causal_framework}), which consists of two complementary branches: the \textit{Personalized Branch} captures local, personalized influences, while the \textit{Global Branch} extracts stable, global traffic patterns.
On the server side, the global patterns and sharable parameters from all clients are aggregated to facilitate global knowledge sharing (Section~\ref{subsec:federated_optimization}).
After receiving the aggregated results from the server, each client combines the two branches to generate the final traffic predictions.

\subsection{Spatial-Temporal Feature Extractor}
\label{subsec:st_feature}
Adaptive adjacency matrix has been widely adopted in spatial-temporal prediction \cite{zhao2019t, li2017diffusion, yu2017spatio, guo2019attention}, as it allows the model to dynamically capture interdependencies among different nodes. 
Motivated by this, we construct an adaptive adjacency matrix
$\mathbf{A}=\mathrm{Softmax}(\mathrm{ReLU}(\mathbf{E}\mathbf{E}^\top))$,
where $\mathbf{E}\in\mathbb{R}^{|\mathcal{V}^{(m)}|\times d}$ is a learnable node embedding matrix. The resulting $\mathbf{A}\in\mathbb{R}^{|\mathcal{V}^{(m)}|\times|\mathcal{V}^{(m)}|}$ is incorporated into graph convolution and further combined with GRU~\cite{cho2014learning} to form an AGR cell~\cite{bai2020adaptive}:

\begin{equation}
\text{GraphConv}(\mathbf{X}, {\mathbf{A}}; \mathbf{W}, \mathbf{b}) = \mathbf{A}\mathbf{X}\mathbf{W} + \mathbf{b},
\end{equation}
where $\mathbf{W}$ and $\mathbf{b}$ are trainable parameters. Given spatial-temporal data $\mathcal{X}\in\mathbb{R}^{|\mathcal{V}^{(m)}|\times T\times F}$, $\text{GraphConv}(\cdot)$ can capture the spatial dependencies among nodes and yield transformed representations.

Then, we extend the GRU as AGR by replacing its linear transformations with adaptive graph convolution operator:

\begin{equation}
\mathbf{z}_t = \sigma\Big(\text{GraphConv}([\mathbf{\mathcal{X}}_t || \mathbf{H}_{t-1}], {\mathbf{A}}; \mathbf{W}_z, \mathbf{b}_z)\Big),
\end{equation}
\begin{equation}
\mathbf{r}_t = \sigma\Big(\text{GraphConv}([\mathbf{\mathcal{X}}_t ||  \mathbf{H}_{t-1}], {\mathbf{A}}; \mathbf{W}_r, \mathbf{b}_r)\Big),
\end{equation}
\begin{equation}
\tilde{\mathbf{H}}_t = \tanh\Big(\text{GraphConv}([\mathbf{\mathcal{X}}_t ||  (\mathbf{r}_t \odot \mathbf{H}_{t-1})], \mathbf{A}; \mathbf{W}_h, \mathbf{b}_h)\Big),
\end{equation}
\begin{equation}
\mathbf{H}_t = \mathbf{z}_t \odot \mathbf{H}_{t-1} + (1-\mathbf{z}_t)\odot \tilde{\mathbf{H}}_t.
\end{equation}
where $\sigma(\cdot)$ denotes the sigmoid function, $\odot$ denotes the Hadamard product, $||$ indicates concatenation. The variables $\mathbf{z}_t$, $\mathbf{r}_t$, and $\tilde{\mathbf{H}}_t$ correspond to the update gate, the reset gate, and the candidate hidden state in the AGR cell, respectively.

For the input traffic sequence $\mathcal{X}$, we employ a spatial-temporal encoder composed of stacked AGR layers. 
The encoder processes the sequence recurrently: at each time step $t$, every AGR layer computes its current hidden state based on the input feature $x_t$ and the previous hidden state $\mathbf{H}_{t-1}$. 
The output sequence from the final layer of the stack, $\mathbf{H} \in \mathbb{R}^{|\mathcal{V}^{(m)}| \times C}$, serves as the comprehensive spatial-temporal representation, where $C$ is the hidden dimension.

This encoder architecture serves as the foundation of our causal representation decoupling framework. 
We implement two structurally identical encoders with unshared parameters to maintain the spatial-temporal representations into two distinct latent components, \ie $\mathbf{S}$ and $\mathbf{D}$.
This design allows each branch to specialize in modeling different facets of the spatial-temporal dynamics.

\subsection{Causal Representation Decoupling Framework}
\label{subsec:causal_framework}
In this subsection, we propose a causal decoupling framework that explicitly separates personalized, dynamic fluctuations from stable, globally shared patterns. Our model features two complementary branches: a personalized branch to capture client-specific influences and a global branch to model common traffic behaviors, providing a comprehensive and disentangled view of the overall traffic state.

\subsubsection{\textbf{Personalized Branch.}}
\begin{figure}[!t]
    \centering
    \includegraphics[width=\linewidth]{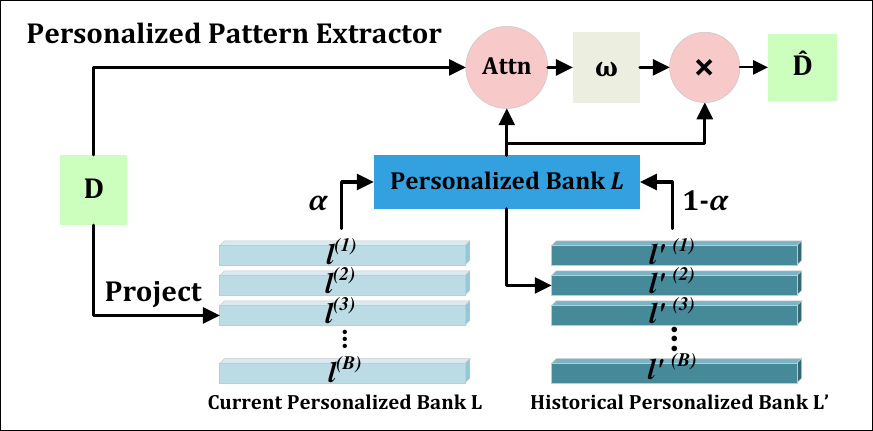}
    \caption{Personalized pattern extractor.}
    \label{fig:DFE}
\end{figure}

The Personalized Branch is designed to capture the unique local traffic behaviors of each individual client through a pattern bank that encodes latent personalized states.
Its core component is a Personalized Pattern Extractor (Figure~\ref{fig:DFE}), which maintains a bank of latent local patterns and adaptively retrieves them to inform the model's predictions. This allows the model to adjust to localized and ever-changing traffic dynamics.

\noindent\textit{\textbf{Personalized Pattern Extractor.}}

The personalized pattern bank is denoted as $\mathbf{L} \in \mathbb{R}^{B \times C}$, where $B$ denotes the number of personalized patterns.
This bank is designed to store and evolve a set of representative local traffic states. To ensure the bank is both stable and adaptive, it is randomly initialized, refined via PCA whitening ~\cite{abdi2010principal}, and then updated with momentum. Given the current traffic representation $\mathbf{D} \in \mathbb{R}^{|\mathcal{V}^{(m)}| \times C}$ from the upstream AGR layers, we first project it to generate a set of current patterns $\mathbf{L}$. The bank $\mathbf{L}$ is then updated as $\mathbf{L} = \alpha \mathbf{L} + (1-\alpha)\mathbf{L}'$.
Here, $\alpha \in [0,1]$ is a momentum coefficient that balances the influence of historical patterns (the existing $\mathbf{L}'$) against new observations ($\mathbf{L}$). 
This allows the bank to preserve long-term context while integrating the most recent traffic dynamics.

With the updated pattern bank, the model must determine which historical patterns are most relevant to the current traffic state. 
To achieve this, we use an attention mechanism. For each feature vector $\mathbf{D}^{(i)}$ in the current representation $\mathbf{D}$, we compute a compatibility score $s^{(i)}$ between $\mathbf{D}^{(i)}$ and every vector $l^{(k)}$ in the pattern bank. 
These scores are calculated using a learnable function, typically a lightweight MLP, and then normalized via a softmax operation to produce attention weights $\omega = \text{Softmax}(s)$.
The refined personalized representation $\hat{\mathbf{D}}$ is then computed as a weighted combination of the bank vectors:
\begin{equation}
    \hat{\mathbf{D}}_i = \sum_{k=1}^{B} \omega^{(i,k)} l^{(k)},
\end{equation}
where $\omega^{(i)}$ denotes the $i$-th row of $\omega$, and $\hat{\mathbf{D}}_i$ is the $i$-th row of the final personalized representation $\hat{\mathbf{D}}$.

The personalized representation is obtained by selectively attending to the most relevant vectors in the personalized bank. Through this attention-based fusion, the model adaptively captures diverse personalized patterns, allowing the learned representations to reflect heterogeneous dynamics under varying traffic contexts. 

\noindent\textit{\textbf{Personalized Predictor.}}
The Personalized Branch generates its prediction by fusing the original dynamic features $\mathbf{D}$ with the refined, context-aware representation $\hat{\mathbf{D}}$.
This integration allows the branch to adaptively model heterogeneous and client-specific traffic patterns. The combined features are then mapped to the output dimension via a a fully connected network $\hat{\mathbf{Y}}^P = \text{MLP}(\mathbf{D} + \hat{\mathbf{D}}).$

\subsubsection{\textbf{Global Branch.}}
The Global Branch is designed to capture stable, invariant traffic patterns that are common across all clients. 
Its primary component, a Global Pattern Extractor, distills these shared features while a Global Predictor generates the final branch output. 
This design ensures that the learned global patterns are robustly disentangled from local variations.

\noindent\textit{\textbf{Global Pattern Extractor.}}
\begin{figure}[!t]
    \centering
    \includegraphics[width=\linewidth]{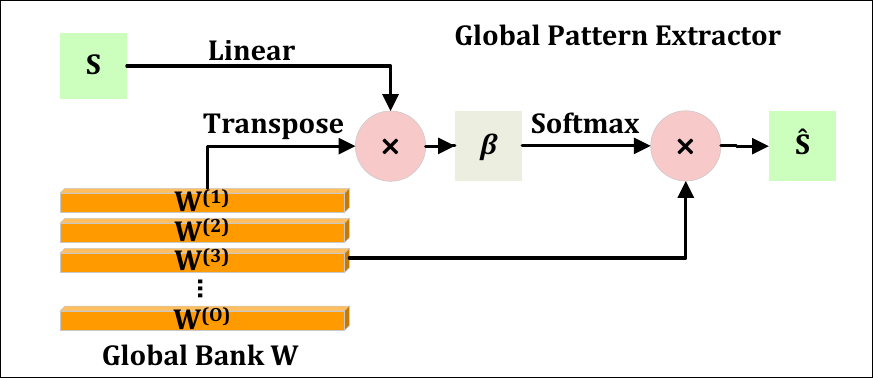} 
    \caption{Global pattern extractor. }
    \label{fig:GPE}
\end{figure}

As illustrated in Figure~\ref{fig:GPE}, the global pattern extractor's main goal is to form a stable global representation without directly sharing features derived from raw client data, thus preserving privacy.
To achieve this, it maintains a learnable global pattern bank $\mathbf{W} \in \mathbb{R}^{O \times C}$, where $O$ denotes the number of global patterns. The bank uses random initialization followed by Xavier ~\cite{glorot2010understanding} and continuously refined during training to encode globally consistent traffic patterns.

Each feature in $\mathbf{S}$ is projected into a query space through a learnable linear mapping, yielding $\overline{\mathbf{S}} = \mathbf{S} \cdot \mathbf{W}_s + \mathbf{b}_s$, where $\mathbf{W}_s$ and $\mathbf{b}_s$ are learnable projection parameters. The resulting queries interact with the pattern bank $\mathbf{W}$ to compute similarity scores, which are normalized via softmax to obtain attention weights $\beta \in \mathbb{R}^{|\mathcal{V}^m| \times O}$. These weights guide the aggregation of shared patterns, producing the final global representation $\hat{\mathbf{S}} \in \mathbb{R}^{|\mathcal{V}^m| \times C}$:
\begin{equation}
    \hat{\mathbf{S}} = \beta \cdot \mathbf{W} = \text{Softmax}(\overline{\mathbf{S}} \cdot \mathbf{W}^\top) \cdot \mathbf{W}.
\end{equation}

\noindent\textit{\textbf{Global Predictor.}}
{The global predictor fuses the refined global representation $\hat{\mathbf{S}}$ with the stable features $\mathbf{S}$ to generate the final prediction. By emphasizing invariant and cross-client dependencies, it captures globally consistent trends that complement the personalized branch. The mapping is implemented through a fully connected network, formulated as: $\hat{\mathbf{Y}}^U = \text{MLP}(\mathbf{S} + \hat{\mathbf{S}}).$

\subsubsection{\textbf{Causal Disentanglement}}
\label{Mutual Information Causal Decoupling}
To uncover the underlying mechanisms driving the traffic states, we adopt the Structural Causal Model (SCM) \cite{neuberg2003causality} to describe the relationships between historical and future traffic states.
Beyond direct effect $\mathcal{X} \to \mathcal{Y}$, we define personal dynamic patterns $\hat{\mathbf{D}}$ that act as latent causal variables influencing both history and future, which can be expressed as $\mathcal{X} \gets \hat{\mathbf{D}} \to \mathcal{Y}$. These patterns capture local and time-varying influences not directly reflected in historical traffic data.

We minimize the mutual information $MI(\mathbf{S},\hat{\mathbf{D}})$ between the global feature $\mathbf{S}$ and the personalized representation $\hat{\mathbf{D}}$. This objective reduces redundancy between the two branches, prevents $\mathbf{S}$ from absorbing personalized dynamics, and prevents $\hat{\mathbf{D}}$ from re-encoding global structure, leading to more disentangled and interpretable components.
Directly enforcing independence between $\mathbf{S}$ and $\hat{\mathbf{D}}$ is intractable since the exact mutual information $MI(\mathbf{S},\hat{\mathbf{D}})$ has no closed form solution. We therefore adopt the Contrastive Log-ratio Upper Bound (CLUB) \cite{cheng2020club}, which provides a sample-based \emph{upper bound} suitable for minimization. CLUB models the conditional distribution $q_\theta(\mathbf{S}|\hat{\mathbf{D}})$ with a Gaussian parameterized by an MLP, allowing the bound to be estimated efficiently from mini-batch samples. Minimizing this upper bound during training enforces causal disentanglement between $\mathbf{S}$ and $\hat{\mathbf{D}}$.
Given minibatch samples $\{(\mathbf{S}_i,\hat{\mathbf{D}}_i)\}_{i=1}^{I}$, CLUB estimates the upper bound as:
\begin{equation}
    \widehat{MI}_{\mathrm{CLUB}}(\mathbf{S},\hat{\mathbf{D}}) = \frac{1}{U} \sum_{i=1}^{U} \Big[ \log q_\delta(\mathbf{S}_i \mid \hat{\mathbf{D}}_i) - \frac{1}{U} \sum_{j=1}^{U} \log q_\delta(\mathbf{S}_j \mid \hat{\mathbf{D}}_i) \Big],
\end{equation}
where $q_\delta(\mathbf{S}|\hat{\mathbf{D}})$ is a variational conditional Gaussian parameterized by a small MLP:
\begin{equation}
    q_\delta(\mathbf{S}|\hat{\mathbf{D}}) = \mathcal{N}\big(\mathbf{S};\, \mu_\delta(\hat{\mathbf{D}}), \mathrm{diag}(\sigma_\delta^2(\hat{\mathbf{D}}))\big),
\end{equation}
\begin{equation}
    \{\mu_\delta(\hat{\mathbf{D}}),\sigma_\delta^2(\hat{\mathbf{D}})\} = \mathrm{MLP}_\delta(\hat{\mathbf{D}}).
\end{equation}
The MI loss is then defined as: 
\begin{equation}
    \mathcal{L}_{{MI}} = \widehat{MI}_{\mathrm{CLUB}}(\mathbf{S},\hat{\mathbf{D}}).
\end{equation}

By minimizing $\mathcal{L}_{{MI}}$, we reduce redundancy between $\mathbf{S}$ and $\hat{\mathbf{D}}$.
Through maintaining these two factors locally and collaboratively, we ensure that the global branch encodes global, invariant traffic patterns while the personalized branch accurately captures dynamic, local traffic behaviors.

% \textbf{P5. Universal-Driven Federated Collaborative Optimization}

\subsection{Federated Optimization Objectives}
\label{subsec:federated_optimization}
Our federated framework enhances traffic prediction collaboratively across multiple clients while preserving data privacy. It employs collaborative pattern sharing to exchange global traffic knowledge through a learnable pattern bank without exposing raw local data, and graph attention fusion to construct graph prototypes via attention, where prototype similarities on the server guide efficient and privacy-preserving parameter aggregation.

\subsubsection{\textbf{Collaborative Pattern Sharing.}}
To enhance knowledge sharing across all participating clients, the framework allows them to collaboratively refine and update the global traffic patterns. Each client uploads its pattern bank $\mathbf{W}^{(m)}$ to the server, where similarity-aware aggregation consolidates globally shared traffic patterns while preserving client-specific adaptations.

For each pattern $\mathbf{W}^{(m)}[j]$ in client $m$'s pattern bank, the server identifies the top-$K$ most similar patterns from other clients’ banks, considering only those whose similarity exceeds a predefined threshold $\tau$.
Formally, for a given pattern $\mathbf{W}^{(m)}[j]$ in client $m$’s bank, the selection set is defined as:
\begin{equation}
\begin{aligned}
    \mathcal{\mathbf{P}}^{(m)}_{j} = \Big\{ (n,k) \;\Big|\; & n \neq m, k \in \text{Top-K}\big( \text{sim}(\mathbf{W}^{(m)}[j], \mathbf{W}^{(n)}) \big), \\
    & \text{sim}(\mathbf{W}^{(m)}[j], \mathbf{W}^{(n)}[k]) > \tau \Big\},
\end{aligned}
\end{equation}
where $\text{sim}(\cdot,\cdot)$ denotes the cosine similarity between two patterns, $n$ indexes other clients, and $k$ indexes the top-$K$ similar patterns in client $n$’s bank that satisfy the threshold.

The aggregation then updates each pattern as:
\begin{equation}
    \mathbf{W}^{(m)}[j] \;\gets\; \frac{1}{|\mathcal{\mathbf{P}}^{(m)}_{j}|} \sum_{(n,k) \in \mathcal{\mathbf{P}}^{(m)}_{j}} \mathbf{W}^{(n)}[k],
\end{equation}
where \(|\mathcal{\mathbf{P}}^{(m)}_{j}|\) is the number of patterns selected for aggregation and is used to normalize the weighted sum.

After aggregation, the updated banks $\mathbf{W}^{(m)}$ are redistributed to the corresponding clients for the next round of local training. This similarity-aware aggregation ensures that shared traffic patterns are aligned across clients while preserving client-specific diversity.

\subsubsection{\textbf{Graph Attention Fusion.}}
\label{Graph Attention Fusion}

To enable the server to measure inter-client similarities and perform structure-aware parameter aggregation, we introduce Graph Attention Fusion. This mechanism enhances collaboration among clients while preserving their inherent structural diversity by representing each client’s local graph characteristics with a prototype.

Each client first extracts node embeddings $\mathbf{h}_v$ from its local model, capturing the structural and spatial dependencies within its traffic network. These embeddings are then aggregated into a graph prototype (GP) using a graph attention mechanism:
\begin{equation}
    \mathbf{s}_v = \mathbf{w}^\top \cdot \tanh(\mathbf{W}_v \cdot \mathbf{h}_v + \mathbf{b}_v),
\end{equation}
where $\mathbf{W}_v$ and $\mathbf{b}_v$ are learnable projection parameters, and $\mathbf{w}$ is a learnable vector that maps the projected features to a scalar importance score. Attention weights $\alpha_v$ are obtained via softmax, and the GP is computed as:
\begin{equation}
    \mathbf{h}_G = \sum_{v=1}^{N} \alpha_v \cdot \mathbf{h}_v, \quad 
    \alpha_v = \frac{\exp(\mathbf{s}_v)}{\sum_{u=1}^{|\mathcal{V}^{(m)}|} \exp(\mathbf{s}_u)}.
\end{equation}

At the server, GPs from all clients are used to compute inter-client similarities, converted into attention weights via a temperature parameter $\epsilon$. Each client’s sharable parameters $\theta_u^{(i,s)}$ are then updated as a similarity-weighted combination of all clients’ parameters:
\begin{equation}
\theta_u^{(i,s)} = \sum_{j=1}^M \frac{\exp(Sim(\mathbf{h}_i,\mathbf{h}_j)/\epsilon)}{\sum_u \exp(Sim(\mathbf{h}_i,\mathbf{h}_u)/\epsilon)} \theta_u^{(j,s)}.
\end{equation}

The server redistributes the updated parameters for the next training round. This iterative process ensures that clients with similar traffic patterns share knowledge effectively while preserving personalized representations.

\subsubsection{\textbf{Federated Optimization.}} 

In federated optimization, the server coordinates multiple clients to collaboratively build a shared model without ever exposing their raw, sensitive data.

\noindent\textit{\textbf{Server-side.}}
In each federated round, the server collects updates from all clients, including the global pattern bank $\mathbf{W}^{(m)}$, graph prototypes $\mathbf{h}_G^{(m)}$, and shared model parameters $\theta^{(m,s)}$. It aggregates similar patterns across clients through collaborative pattern sharing and updates the shared parameters via graph attention fusion guided by the graph prototypes. The refined parameters and pattern banks are then redistributed to clients.

\noindent\textit{\textbf{Client-side.}}
Each client receives the updated shared parameters $\bar{\theta}^{(m,s)}$ and global pattern bank $\bar{\mathbf{W}}^{(m)}$ from the server and performs local training on both global and personalized branches. Let $\hat{\mathbf{Y}}^U$ and $\hat{\mathbf{Y}}^P$ denote the outputs from the global and personalized branches, fused as $\hat{\mathbf{Y}} = \hat{\mathbf{Y}}^U + \hat{\mathbf{Y}}^P$.

The local objective combines prediction supervision via MAE loss $\mathcal{L}_{\text{pred}}$ and mutual information loss $\mathcal{L}_{MI}$ to disentangle global and personalized features, where $\lambda$ balances the two terms:
\begin{equation}
    \mathcal{L} = \mathcal{L}_{\text{pred}} + \lambda \mathcal{L}_{MI}.
\end{equation}

During training pipeline, only global branch parameters $\theta^{(m,s)}$, graph prototypes $\mathbf{h}_G^{(m)}$, and the global pattern bank $\mathbf{W}^{(m)}$ are uploaded to the server; personalized parameters $\theta^{(m,l)}$ remain local, preserving client-specific information.

%% file: sections/5_Experiments.tex
\section{Experiments}
This section presents the experimental results, including comparisons with state-of-the-art baselines, evaluation under different federated client configurations, ablation studies on core model components, and analysis of key hyper-parameters.

\subsection{Datasets}

Our experiments are conducted on four widely used traffic benchmarks: METR-LA, PEMS-BAY \cite{li2017diffusion}, PEMS03, and PEMS04 \cite{song2020spatial}. METR-LA and PEMS-BAY record traffic speed from loop detectors in Los Angeles and the Bay Area, while PEMS03 and PEMS04 contain traffic flow measurements from the California PeMS system. All datasets have 5-minute intervals (288 per day). For fair comparison, each dataset is split into 60\% training, 20\% validation, and 20\% testing. Detailed statistics are provided in Table~\ref{tab:dataset}.

\begin{table}[!t]
\centering
\caption{Datasets statistics.}
\begin{tabular}{cccc}
\hline
\textbf{Dataset} & \textbf{\#Sensors} & \textbf{\#Timesteps} & \textbf{Time Range} \\
\hline
METR-LA  & 207 & 34,272 & 03/2012 -- 06/2012 \\
PEMS-BAY & 325 & 52,116 & 01/2017 -- 06/2017 \\
PEMS03   & 358 & 26,208 & 09/2018 -- 11/2018 \\
PEMS04   & 307 & 16,992 & 01/2018 -- 02/2018 \\
\hline
\end{tabular}
\label{tab:dataset}
\end{table}

\begin{table*}[!t]
\centering
\small
\caption{Performance comparison of different methods. 
``(a)'' means centralized traffic prediction methods, 
``(b)'' means federated traffic prediction methods, 
and ``(ours)'' means our proposed method \name. 
``Time/r'' represents the average time of federation rounds. 
Within each method group, the best results are \textbf{bolded} and the second-best results are \underline{underlined}.}
\renewcommand{\arraystretch}{0.9} 
\resizebox{\textwidth}{!}{%
\begin{tabular}{c c c c c c c c c c c c c c c}
\toprule
 & \textbf{Dataset} & \multicolumn{3}{c}{\textbf{METR-LA}} & \multicolumn{3}{c}{\textbf{PEMS-BAY}} & \multicolumn{3}{c}{\textbf{PEMS03}} & \multicolumn{3}{c}{\textbf{PEMS04}} & Time/r \\ 
\cmidrule(lr){3-5} \cmidrule(lr){6-8} \cmidrule(lr){9-11} \cmidrule(lr){12-14} 
 & \textbf{Metric} & MAE & RMSE & MAPE(\%) & MAE & RMSE & MAPE(\%) & MAE & RMSE & MAPE(\%) & MAE & RMSE & MAPE(\%) & (s) \\
\midrule
\multirow[c]{2}{*}{(a)} & GWNet  & \textbf{3.05} & \textbf{6.20} & \textbf{8.29} & \textbf{1.59} & \textbf{3.66} & \textbf{3.54} & \textbf{14.82} & \textbf{25.24} & 16.16 & \textbf{19.16} & \textbf{30.46} & 13.26 & - \\
 & AGCRN  & \underline{3.19} & \underline{6.46} & \underline{9.00} & \underline{1.64} & \underline{3.77} & \underline{3.71} & \underline{15.89} & 28.12 & \underline{15.38} & \underline{19.83} & 32.26 & \underline{12.97} & - \\
\midrule
\multirow[c]{1}{*}{(ours)} & \textbf{\name} & 3.55 & 7.01 & 9.66 & 1.75 & 3.90 & 3.93 & 16.10 & \underline{25.92} & \textbf{13.43} & 20.66 & \underline{32.07} & \textbf{12.43} & - \\
\midrule
\multirow[c]{5}{*}{(b)} 
 & FedAvg  & 4.14 & 9.48 & 13.56 & \underline{1.84} & 4.09 & 4.18 & 18.71 & 29.48 & 15.92 & 26.50 & 43.37 & 16.55 & 214.61 \\
 & FedProx & 3.81 & 7.53 & 10.80 & 2.15 & 5.97 & 4.83 & 18.52 & 33.77 & 16.12 & 26.20 & 40.96 & 16.78 & 220.99 \\
 & STDN  & \underline{3.57} & 7.31 & 11.06 & 1.86 & 4.43 & 4.17 & 18.59 & 30.21 & 15.38 & 21.77 & 33.21 & \underline{13.39} & 101.80  \\
 & FedGTP  & 4.86 & 10.51 & 13.12 & \textbf{1.75}  & \underline{3.98}  & \underline{3.95}  & \underline{17.31} & 27.98 & 17.97 & 22.15 & 34.17 & 16.31 & 1,758.45 \\
 & pFedCTP & 3.76 & \underline{7.07} & \underline{10.44} & 2.20 & 4.49 & 5.04 & 17.58 & \underline{26.72} & \underline{14.29} & \underline{21.53} & \underline{33.16} & 14.35 & 138.12 \\
\midrule
\multirow[c]{1}{*}{(ours)} & \textbf{\name} & \textbf{3.55} & \textbf{7.01} & \textbf{9.66} & \textbf{1.75} & \textbf{3.90} & \textbf{3.93} & \textbf{16.10} & \textbf{25.92} & \textbf{13.43} & \textbf{20.66} & \textbf{32.07} & \textbf{12.43} & 253.69 \\
\midrule
\multirow[c]{1}{*}{} & \textbf{Impr$\uparrow$} & \textbf{$+$0.56\%} & \textbf{$+$0.85\%} & \textbf{$+$7.47\%} & \textbf{$+$0.49\%} & \textbf{$+$0.21\%} & \textbf{$+$0.51\%} & \textbf{$+$6.99\%} & \textbf{$+$2.99\%} & \textbf{$+$6.02\%} & \textbf{$+$4.04\%} & \textbf{$+$3.29\%} & \textbf{$+$7.17\%} & - \\
\bottomrule
\end{tabular}%
}
\label{tab:baseline_comparison_grouped}
\end{table*}

\begin{table}[!t]
\centering
\small
\caption{Performance comparison under different numbers of clients. 
``\#Clients'' denotes the number of clients. 
}
\renewcommand{\arraystretch}{0.8}
\resizebox{0.9\linewidth}{!}{%
\begin{tabular}{c c c c c c}
\toprule
\textbf{\# Clients} & \textbf{Metric} & \textbf{FedGTP} & \textbf{pFedCTP} & \textbf{\name} & \textbf{Impr$\uparrow$}\\
\midrule
\multirow{3}{*}{4} 
 & MAE & 4.86 & \underline{3.76} & \textbf{3.55} & \textbf{$+$5.59\%} \\
 & RMSE & 10.51 & \underline{7.07} & \textbf{7.01} & \textbf{$+$0.85\%} \\
 & MAPE(\%) & 13.12 & \underline{10.44} & \textbf{9.66} & \textbf{$+$7.47\%}  \\
\midrule
\multirow{3}{*}{8} 
 & MAE & 5.01 & \underline{3.86} & \textbf{3.59} & \textbf{$+$6.99\%}  \\
 & RMSE & 12.45 & \underline{6.81} & \textbf{6.79} & \textbf{$+$0.29\%} \\
 & MAPE(\%) & 12.76 & \underline{10.54} & \textbf{9.84} & \textbf{$+$6.64\%} \\
\midrule
\multirow{3}{*}{16} 
 & MAE & 4.81 & \underline{3.74} & \textbf{3.60} & \textbf{$+$3.74\%} \\
 & RMSE & 12.10 & \textbf{6.93} & \underline{6.96} & - \\
 & MAPE(\%) & 12.22 & \underline{10.36} & \textbf{9.95} & \textbf{$+$3.96\%}\\
\midrule
\multirow{3}{*}{32} 
 & MAE & 5.10 & \underline{3.87} & \textbf{3.64} & \textbf{$+$5.94\%} \\
 & RMSE & 12.33 & \underline{7.14} & \textbf{7.08} & \textbf{$+$0.84\%} \\
 & MAPE(\%) & 12.72 & \underline{10.60} & \textbf{10.17} & \textbf{$+$4.06\%} \\
\bottomrule
\end{tabular}%
}
\label{tab:client_number_label}
\end{table}

\begin{figure*}[!t]
    \centering
    \includegraphics[width=\linewidth]{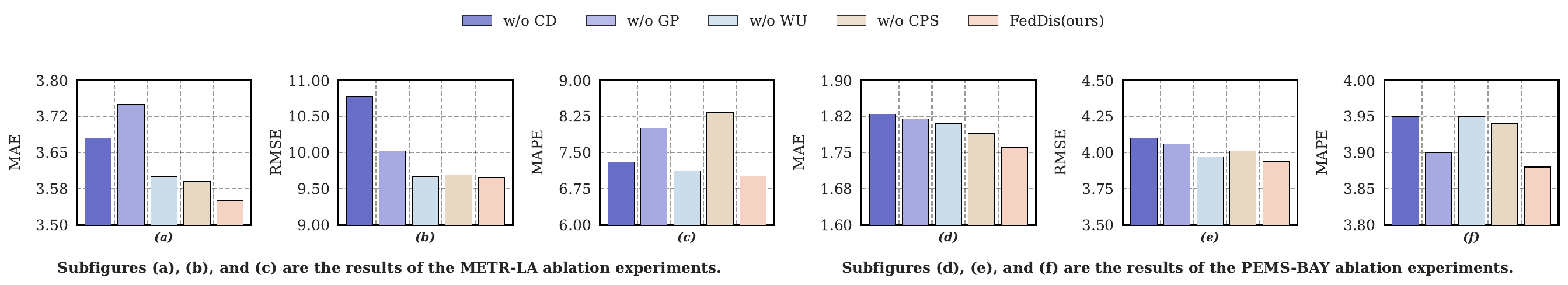} 
    \caption{Ablation study.}
    \label{fig:ablation}
\end{figure*}

\subsection{Baselines}
To evaluate the proposed model, we compare it with two groups of baselines using MAE, RMSE, and MAPE:

(1) \textbf{Centralized traffic prediction methods:} Serve as upper-bound references, assuming all data are available centrally. \textbf{GWNet} \cite{wu2019graph}: Captures multiscale spatial and temporal dependencies via graph wavelet transforms. \textbf{AGCRN} \cite{bai2020adaptive}: Learns dynamic spatial correlations with adaptive graph convolution and models temporal patterns using recurrent networks.  

(2) \textbf{Federated traffic prediction methods:} Includes general federated optimization strategies and recent spatial-temporal traffic prediction approaches. \textbf{FedAvg} \cite{mcmahan2017communication}: Averages client model updates. \textbf{FedProx} \cite{li2020federated}: Adds a proximal term to mitigate heterogeneity across clients. \textbf{STDN} \cite{cao2025spatiotemporal}: Trend-seasonality disentanglement-based and transformer-based model with federated setting. \textbf{FedGTP} \cite{yang2024fedgtp}: Models inter-client spatial dependencies with a graph-based approach. \textbf{pFedCTP} \cite{zhang2024personalized}: Decouples spatial and temporal components for personalized cross-city traffic prediction.

\subsection{Experimental Setups}
Our framework is implemented in Python 3.12.3 with PyTorch 2.7.0, and experiments are conducted on NVIDIA RTX 3090 GPUs. The look-back window and prediction horizon are set to 12, with 4 clients by default. We train the model using Adam with an initial learning rate of 0.005 and a batch size of 64. The momentum coefficient $\alpha$ for personalized bank updates is set to 0.5, and the similarity threshold $\tau$ for collaborative pattern sharing is 0.3.
We implement baselines and tune key hyper-parameters to attain the optimal performance.
Federated training runs for 100 communication rounds, with each client performing 1 local epoch per round.

\subsection{Overall Performance}
As shown in Table~\ref{tab:baseline_comparison_grouped}, 
(1)Centralized models are strong baselines due to their efficient use of global data, while federated methods usually underperform because of data decentralization. Although our model does not consistently surpass centralized approaches, it achieves competitive performance under privacy constraints and shows advantages on specific metrics, such as lower MAPE on PEMS03 and PEMS04. These gains stem from the synergy of causal representation decoupling and cross-client collaboration, where disentangling global and client-specific factors produces sharable, interference-resistant representations, and similarity-aware aggregation with graph attention further improves global knowledge fusion and consistency under privacy constraints.
(2) FedAvg and FedProx are implemented as federated variants of our backbone for fair comparison. They perform competitively on METR-LA and PEMS-BAY, confirming the strong representational capacity of our backbone, but degrade on sparser datasets (PEMS03/04) due to limited handling of client heterogeneity. In contrast, our method explicitly models structure-aware and pattern-level heterogeneity, yielding consistently lower prediction errors across all datasets. 
(3) Within the federated traffic prediction methods, our method consistently achieves the lowest or near-lowest error across datasets, demonstrating stable and accurate spatial-temporal predictions. Other methods also perform competitively, showing that federated designs can effectively share knowledge across clients while capturing spatial-temporal dependencies under heterogeneous data. For instance, it reduces MAE by 6.99\% over FedGTP and 6.02\% over pFedCTP on PEMS03. These consistent gains confirm that our approach further improves robustness and prediction accuracy by explicitly modeling both personalized and global traffic patterns.
(4) In terms of computational efficiency, our method maintains competitive costs among federated traffic prediction baselines. \name required 253.69s per round, comparable to FedAvg and FedProx, and far more efficient than FedGTP (1,758.45s), indicating that the introduced collaboration mechanisms add only modest overhead. The \emph{similarity-aware pattern aggregation} and \emph{graph attention fusion} modules operate on compact pattern banks and graph prototypes rather than full model parameters, ensuring lightweight computation. As a result, the model effectively leverages shared knowledge without substantially increasing per-round computation or communication, achieving a favorable trade-off between predictive performance and efficiency in large-scale federated traffic prediction scenarios.

\subsection{Expandability Analysis}
A critical test for real-world federated learning systems is their ability to maintain performance as the network scales. To simulate this, we evaluate model robustness by progressively increasing the number of clients ($N=4,8,16,32$) on the METR-LA dataset.

The results, presented in Table~\ref{tab:client_number_label}, reveal key advantages of our approach. 
While baseline methods exhibit notable performance degradation as client numbers grow, \name demonstrates remarkable resilience, maintaining its state-of-the-art accuracy across all scales. This superior scalability is a direct outcome of our model's fundamental design: by explicitly disentangling personalized dynamics from global patterns, \name is not overwhelmed by the increasing client heterogeneity that causes other entangled models to falter. This architectural choice ensures that the global model learns only from stable, shared knowledge, preserving its integrity regardless of network size. These findings highlight a key advantage of \name—its suitability for dynamic, large-scale federated systems where robustness under increasing complexity is essential.

\begin{figure*}[!t]
    \centering
    \includegraphics[width=\textwidth]{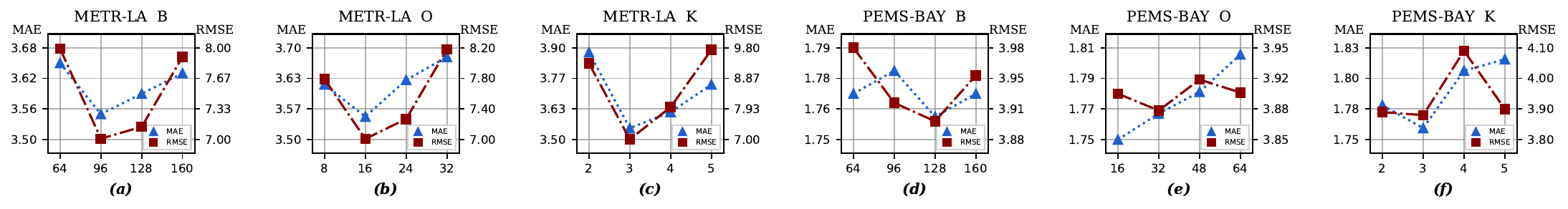} % 自适应当前栏宽
    \caption{Hyper-parameter analysis of the number of personalized patterns $B$, the number of global patterns $O$, and the Top-$K$ patterns on METR-LA and PEMS-BAY.}
    \label{fig: hyperparameters-PEMS&METR}
\end{figure*}

\subsection{Ablation Study}
To assess the contribution of each component, we conduct ablation studies with the following variants: (1) \textbf{w/o CD} removes the causal disentanglement module, including the personalized pattern extractor and mutual information minimization; (2) \textbf{w/o GP} replaces graph attention fusion with standard FedAvg; (3) \textbf{w/o WU} disables the uploading of the global bank, making each client update it locally; and (4) \textbf{w/o CPS} removes the collaborative pattern sharing module and directly averages the global bank on the server before being distributed back to clients. 

The quantitative results are presented in Figure~\ref{fig:ablation}, subfigures (a-c) show the ablation results on the METR-LA dataset, while subfigures (d-f) present the corresponding results on the PEMS-BAY dataset. From these figures, we observe that all components contribute positively to performance.
Specifically, removing \textbf{CD} or \textbf{GP} results in the most notable degradation across all metrics, showing that causal disentanglement and graph-aware aggregation are fundamental for capturing heterogeneous spatial-temporal dependencies. 
The absence of \textbf{WU} leads to a moderate decline, indicating that uploading the global bank mainly facilitates cross-client knowledge transfer and global pattern refinement. 
Furthermore, the variant without \textbf{CPS} shows relatively larger RMSE fluctuations on both datasets, suggesting that the collaborative pattern sharing mechanism is crucial for enhancing similarity-aware knowledge exchange among clients while preventing the over-smoothing of diverse traffic patterns. This demonstrates that CPS effectively strengthens global collaboration without compromising representational diversity.

\subsection{Hyper-Parameters Analysis}
In this subsection, we analyze the sensitivity of {{\name}} to key hyper-parameters. Specifically, $O$ is set to $\{8,16,24,32\}$ for METR-LA and $\{16,32,48,64\}$ for PEMS-BAY, $B$ is varied within $\{64,96,128,160\}$, and Top-$K$ is tested with values $\{2,3,4,5\}$. The results are reported on METR-LA and PEMS-BAY datasets, as shown in Figure~\ref{fig: hyperparameters-PEMS&METR}.

From the results, several conclusions can be made: (1) The performance improves first and then degrades as $O$ increases. This suggests that too few universal patterns cannot fully capture the shared knowledge, while too many may introduce redundancy and noise. 
(2) The smaller dataset favors a moderate number of patterns, while the larger dataset requires more. This indicates that larger-scale datasets tend to exhibit more complex dynamics, demanding richer patterns, whereas moderate settings are sufficient for smaller datasets. 
(3) Using a small number of selected patterns yields the best results across datasets. A very small $K$ may overlook useful knowledge, while a large $K$ may introduce mismatched patterns that harm prediction. Therefore, keeping $K$ small ensures more precise and effective knowledge aggregation.

%% file: sections/3_RelatedWorks.tex
\section{Related Work}
\textbf{Spatial-Temporal Prediction.} Spatial-temporal prediction methods aim to effectively capture both temporal dynamics and spatial dependencies~\cite{chen2020multi, geng2019spatiotemporal, yan2018spatial, yu2017spatio}. 
Early methods mainly employed RNNs for temporal modeling~\cite{shi2015convolutional,cai2023forecasting,jiang2021deepcrowd}, often combined with CNNs to extract spatial features~\cite{zhang2017deep}. 
To model real-world topologies, GNNs have emerged as the predominant tool for modeling non-Euclidean spatial correlations. 
Initial GNN-based models like DCRNN~\cite{li2017diffusion} and STGCN~\cite{han2020stgcn} operated on predefined graphs, while later works adaptively learned graph structures to capture hidden inter-node dependencies~\cite{wu2019graph, bai2020adaptive}. 
Parallel efforts leverage attention mechanisms to model time-varying spatial relationships~\cite{zheng2020gman,fang2023spatio}.

Despite achieving strong accuracy, centralized models face limitations due to privacy and data governance. Our proposed \name addresses these challenges by effectively modeling complex spatial-temporal dependencies while preserving locally stored private data, balancing predictive performance with privacy.

\textbf{Federated Spatial-Temporal Prediction. }
FL naturally addresses distributed data and privacy constraints, allowing clients to train models locally and share only parameters~\cite{mcmahan2017communication}. FL has been increasingly applied to spatial-temporal prediction tasks \cite{fu2022federated, liu2024federated}. For instance, FedGRU~\cite{liu2020privacy} integrates FL with gated recurrent units, leveraging ensemble clustering to capture traffic correlations, CNFGNN~\cite{meng2021cross} collects local temporal embeddings from clients and applies GNNs to generate spatial embeddings, which are then returned to the respective clients for forecasting, while FedGTP~\cite{yang2024fedgtp} uses privacy-preserving mechanisms to exploit inter-client spatial dependencies. To handle data heterogeneity, FedTPS~\cite{zhou2024traffic} extracts low-frequency traffic patterns for global sharing, and pFedCTP~\cite{zhang2024personalized} transfers knowledge from data-rich to data-scarce cities for personalized cross-city prediction.

However, these methods largely fail to disentangle local dynamic patterns from global stable patterns in federated spatial-temporal data. To address this, we explicitly model these two complementary components and propose, to the best of our knowledge, the first causal disentanglement framework for federated traffic prediction.

%% file: sections/6_Conclusion.tex
\section{Conclusion}
Accurate traffic prediction is crucial for intelligent transportation systems, but it is challenging under federated settings where data are distributed and privacy must be preserved. In this context, we first identify the long-overlooked entanglement problem between personalized local patterns and global stable patterns, which existing methods fail to address effectively. To tackle this, we propose {\name}, a dual-branch causal disentanglement framework that leverages mutual information minimization, collaborative pattern sharing, and graph attention fusion to adaptively decouple and integrate global and personalized traffic knowledge. Extensive experiments on four real-world traffic datasets demonstrate that {\name} consistently outperforms existing federated baselines and approaches the performance of centralized models, highlighting its effectiveness in robust, privacy-preserving traffic prediction.

%% file: sections/7_Acknowledgements.tex
\section{Acknowledgments}
ZZ is supported by the China Postdoctoral Science Foundation (2025M771587), the Open Funding Programs of State Key Laboratory of AI Safety and the Key R\&D Program of Jilin Province (20260205050GH). JH is supported by the National Key R\&D Program (2024YFB3310200).

%% file: sections/9_Appendix.tex
\section{Technical Appendix}

\subsection{Framework Illustration}
Our proposed federated framework consists of two core components: a server-side optimization procedure and a client-side local training pipeline.

\begin{algorithm}[!b]
\caption{Server-side Federated Optimization}
\label{alg:server}
% ... (Algorithm 1 code) ...
\raggedright
\begin{algorithmic}[1]
\For{federated round $=1$ to $R_f$}
    \State Receive $\{\mathbf{W}^{(m)}, \mathbf{h}_G^{(m)}, \theta^{(m,s)}\}$ from all clients.
    \State \textbf{Collaborative Pattern Sharing:}
    \For{each client $m$}
        \For{each pattern $\mathbf{W}^{(m)}[j]$}
            \State Select each client Top-$K$ patterns
            \State with similarity $>\tau$.
            \If{non-empty set}
                \State $\bar{\mathbf{W}}^{(m)}[j] \gets \text{Weighted Aggregation}$.
            \EndIf
        \EndFor
    \EndFor
    \State \textbf{Graph Attention Fusion:}
    \For{each client GP $(\mathbf{h}_G^{(i)},\mathbf{h}_G^{(j)})$}
        \State $Sim_{ij} \gets \text{sim}(\mathbf{h}_G^{(i)}, \mathbf{h}_G^{(j)})$.
    \EndFor
    \For{each client $i$}
        \State $\bar{\theta}^{(i,s)} \gets \sum_{j=1}^M \text{softmax}_j\big(Sim_{ij}/\tau\big) \cdot \theta^{(j,s)}$.
    \EndFor
    \State Send updated $\{\bar{\mathbf{W}}^{(m)}, \bar{\theta}^{(m,s)}\}$ back to each client.
\EndFor
\end{algorithmic}
\end{algorithm}

\noindent\textit{\textbf{Server-side.}}
We present the pipeline of the server-side federated optimization in Algorithm \ref{alg:server}.
The server coordinates federated learning by aggregating global patterns and shared parameters while keeping client-specific parameters private. At the start of each federated round (line 2), it collects global pattern banks $\mathbf{W}^{(m)}$, graph prototypes $\mathbf{h}_G^{(m)}$, and shared parameters $\theta^{(m,s)}$ from all clients. In \textit{Collaborative Pattern Sharing} (lines 4--9), the server selects the Top-$K$ most similar patterns exceeding threshold $\tau$ for each client and their patterns, updating the global pattern bank via weighted aggregation. Next, in \textit{Graph Attention Fusion} (lines 13--19), it computes pairwise similarities between client graph prototypes (lines 14-16) and updates shared parameters using attention-weighted aggregation (lines 17--19). Finally, the updated patterns and parameters are sent back to clients (line 20).

\begin{algorithm}[b]
\caption{Client-side Local Training (Client $m$)}
\label{alg:client}
% ... (Algorithm 2 code) ...
\raggedright
	{\bf Input}:  Historical Traffic Observations $\mathcal{X}^{(m)}$; Rounds $R_f, R_l$; Hyperparams $\lambda$.
 \\
	{\bf Output}: Local Prediction  $\mathcal{Y}^{(m)}$. \\
\begin{algorithmic}[1]
\State \textbf{Initialize:} Model Parameters $\theta^{(m)}$, Global Pattern Bank $\mathbf{W}^{(m)}$, Personalized Pattern Bank $\mathbf{L}^{(m)}$.
\For{federated round $=1$ to $R_f$}
    \State Receive updated $\{\bar{\mathbf{W}}^{(m)}, \bar{\theta}^{(m,s)}\}$ from server.
    \For{local round $=1$ to $R_l$}
        \State \textit{// Forward Pass}
        \State $\mathbf{S} \gets \text{Spatial-Temporal Feature}(\mathcal{X}^{(m)}, \mathbf{A}^{(m)};\bar{\theta}^{(m,s)})$ 
        \State $\mathbf{D} \gets \text{Spatial-Temporal Feature}(\mathcal{X}^{(m)}, \mathbf{A}^{(m)};\theta^{(m,l)})$
        \State $\hat{\mathbf{D}} \gets \text{Dynamic Factor Extractor}(\mathbf{D},\mathbf{L}^{(m)})$
        \State $\text{MI Causal Decoupling}(\mathbf{S},\hat{\mathbf{D}})$
        \State $\hat{\mathbf{S}} \gets \text{Universal Pattern Extractor}(\mathbf{S},\bar{\mathbf{W}}^{(m)})$        
        \State $\hat{\mathbf{Y}} \gets \text{Predictor}(\mathbf{S},\hat{\mathbf{S}}; \mathbf{D},\hat{\mathbf{D}})$
        \State
        \State \textit{// Loss Calculation and Backward Pass}
        \State Compute loss $\mathcal{L} \gets \mathcal{L}_{\text{pred}}(\hat{\mathbf{Y}}^{(m)}, \mathbf{Y}^{(m)}) + \lambda \mathcal{L}_{MI}$
        \State Update $\bar{\theta}^{(m,s)}, \theta^{(m,l)}, \bar{\mathbf{W}}^{(m)}$ $\gets$ gradient descent on $\mathcal{L}$.
    \EndFor
    \State
    \State \textit{// Prepare for Upload}
    \State Generate graph prototype $\mathbf{h}_G^{(m)}$.
    \State Upload $\{\mathbf{W}^{(m)}, \mathbf{h}_G^{(m)}, \theta^{(m,s)}\}$ to server.
\EndFor
\end{algorithmic}
\end{algorithm}

\subsection{Hyper-Parameters Analysis}
\begin{figure*}[!t]
    \centering
    \includegraphics[width=\textwidth]{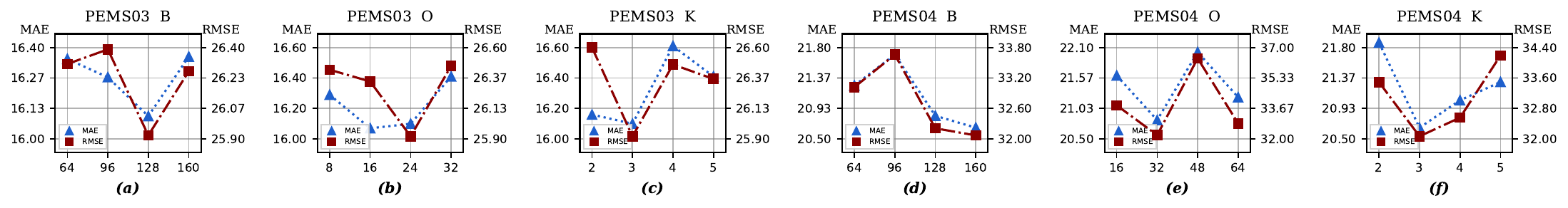} % 自适应当前栏宽
    \caption{Hyper-parameter analysis of the number of personalized patterns $B$, the number of global patterns $O$, and the Top-$K$ patterns on PEMS03 and PEMS04.}
    \label{fig: hyperparameters-pems03&pems04}
\end{figure*}

\noindent\textit{\textbf{Client-side.}}
The pipeline for client-side local training is detailed in Algorithm \ref{alg:client}. Each client performs local training with a dual-branch model to capture both global patterns and personalized dynamics. At the beginning of each federated round, the client receives the updated global pattern bank $\bar{\mathbf{W}}^{(m)}$ and shared parameters $\bar{\theta}^{(m,s)}$ from the server (line 3). In the forward pass, the \textit{global branch} generates static spatial-temporal features $\mathbf{S}$ (line 6), while the \textit{personalized branch} extracts local features $\mathbf{D}$ (line 7). These features are further processed through the \textit{dynamic pattern extractor} to obtain $\hat{\mathbf{D}}$ from the personalized pattern bank $\mathbf{L}^{(m)}$ (line 8), and the \textit{universal pattern extractor} to obtain $\hat{\mathbf{S}}$ from the global pattern bank (line 10). To reduce entanglement between global and personalized information, \textit{MI causal decoupling} is applied between $\mathbf{S}$ and $\hat{\mathbf{D}}$ (line 9). Finally, the predictor combines $\mathbf{S}, \hat{\mathbf{S}}, \mathbf{D}, \hat{\mathbf{D}}$ to produce the local prediction $\hat{\mathbf{Y}}$ (line 11).

After the forward pass, the client calculates a composite loss $\mathcal{L}$ including prediction loss and mutual information loss (line 14). Using this loss, it updates both shared parameters $\bar{\theta}^{(m,s)}$, local private parameters $\theta^{(m,l)}$, and global patterns $\bar{\mathbf{W}}^{(m)}$ through gradient descent (line 15). Once all local rounds are completed, the client generates a new graph prototype $\mathbf{h}_G^{(m)}$ reflecting the current client state (line 19) and uploads the updated global components $\{\bar{\mathbf{W}}^{(m)}, \mathbf{h}_G^{(m)}, \theta^{(m,s)}\}$ to the server (line 20). This design allows each client to retain personalized knowledge while benefiting from collaborative pattern sharing and attention-based aggregation coordinated by the server.

 To further assess the robustness of \name, we conduct hyper-parameter sensitivity analysis on PEMS03 and PEMS04 datasets. We focus on three critical hyper-parameters: the number of universal patterns $O$, the number of dynamic patterns $B$, and the Top-$K$ patterns selected during collaborative pattern sharing. Specifically, for PEMS03, $O$ is set to $\{8,16,24,32\}$, $B$ to $\{64,96,128,160\}$, and Top-$K$ to $\{2,3,4,5\}$. For PEMS04, $O$ is set to $\{16,32,48,64\}$, with the same ranges for $B$ and Top-$K$. The results are reported in Figure~\ref{fig: hyperparameters-pems03&pems04}.

We observe that the model performance first improves and then declines as the number of universal patterns $O$ increases. For the larger PEMS03 dataset, the best MAE and RMSE are obtained at $O=24$, indicating that a moderate number of patterns is sufficient to capture the complex dynamics without introducing redundancy. For the smaller PEMS04 dataset, performance peaks at $O=32$, suggesting that even smaller datasets may require a moderate number of patterns to fully represent shared traffic characteristics.

Regarding the number of dynamic patterns $B$, moderate values generally yield the best results. For PEMS03, $B=128$ achieves the lowest MAE and RMSE, while for PEMS04, $B=160$ slightly improves MAPE and RMSE. Too few dynamic patterns may underfit personalized traffic dynamics, whereas too many can introduce noise.

For Top-$K$ selection, small values consistently lead to better performance across both datasets. Selecting only the most relevant patterns ensures precise knowledge aggregation, whereas larger $K$ values may include mismatched patterns and degrade prediction accuracy. Specifically, $K=3$ achieves the best balance for both PEMS03 and PEMS04. Overall, these results demonstrate that \name is robust to hyper-parameter variations when proper moderate values are chosen.

\subsection{Initialization Study.}
We further investigate the impact of different initialization strategies for the pattern bank. Table~\ref{tab:init_strategy} compares Random, Xavier ~\cite{glorot2010understanding}, and Kaiming ~\cite{he2015delving} initialization with our proposed strategy on the PEMS04 dataset. As shown in Table 4, more appropriate initialization schemes consistently yield better performance, and \name achieves the best results across all evaluation metrics. Compared with standard initialization methods, \name provides a more informative starting point for pattern learning and refinement, enabling the model to capture more discriminative spatial-temporal patterns. These results demonstrate that, beyond architectural design, proper initialization and pattern refinement strategies are critical for achieving optimal performance.

\begin{table}[!t]
\centering
\small
\renewcommand{\arraystretch}{0.8}
\setlength{\tabcolsep}{12pt}
\caption{Performance comparison of different initialization strategies on the PEMS04 dataset.}
\label{tab:init_strategy}
\resizebox{0.9\linewidth}{!}{%
\begin{tabular}{lccc}
\toprule
\textbf{Method} & MAE & RMSE & MAPE(\%) \\
\midrule
Random   & 22.01 & 33.62 & 13.88 \\
Xavier  & 21.70 & 33.26 & 13.65 \\
Kaiming & 20.96 & 32.08 & 13.70 \\
\textbf{FedDis} & \textbf{20.66} & \textbf{32.07} & \textbf{12.43} \\
\bottomrule
\end{tabular}}
\end{table}